# A multi-cohort study on prediction of acute brain dysfunction states using selective state space models

**Brandon Silva [1,3]\*, Miguel Contreras [1,3]\*, Sabyasachi Bandyopadhyay [1,3], Yuanfang Ren [2,3], Ziyuan Guan [2,3], Jeremy Balch, Kia Khezeli [1,3], Tezcan Ozrazgat Baslanti [2,3], Ben Shickel, Azra Bihorac [2,3], Parisa Rashidi [1,3]\*\***

[1]Department of Biomedical Engineering, University of Florida, Gainesville, FL USA

[2]Department of Medicine, University of Florida, Gainesville, FL USA

[3]Intelligent Critical Care Center (IC3), University of Florida, Gainesville, FL USA

\* Denotes equal contribution

\*\* Correspondence:
Parisa Rashidi
parisa.rashidi@ufl.edu

**Abstract**

Assessing acute brain dysfunction (ABD), including delirium and coma in the intensive care unit (ICU), is a critical challenge due to its prevalence and severe implications for patient outcomes. Current diagnostic methods rely on infrequent clinical observations, which can only determine a patient's ABD status after onset. Furthermore, manual observations made by overburdened ICU nurses can increase their fatigue and reduce overall care quality in ICUs. Our research attempts to solve these problems by harnessing the wealth of Electronic Health Records (EHR) data to develop data-driven automated methods for ABD prediction in critically ill patients in the ICU. While a number of predictive models have been introduced in the past, existing models solely predict a single state (e.g., either delirium or coma), require at least 24 hours of observation data to make predictions, do not dynamically predict fluctuating ABD conditions during ICU stay (typically a one-time prediction), and use small sample size, proprietary single-hospital datasets. Our research fills these gaps in the existing literature by dynamically predicting delirium, coma, and mortality for 12-hour intervals throughout an ICU stay and validating on two public datasets. Our research also introduces the concept of dynamically predicting critical transitions from non-ABD to ABD and between different ABD states in real time, which could be clinically more informative for the hospital staff. We compared the predictive performance of two state-of-the-art neural network models, the MAMBA selective state space model and a custom Transformer model, and additionally compared their performance to two traditional machine learning models. Using the MAMBA selective state space model architecture (the best-performing model), we achieved a mean area under the receiving operator characteristic curve (AUROC) of **0.95** on outcome prediction of ABD for 12-hour intervals. Further, the model achieves a mean AUROC of **0.79** when predicting a transition between ABD states. Our study uses a curated dataset from the University of Florida Health Shands Hospital for internal validation and two publicly available datasets, MIMIC-IV and eICU, for external validation, demonstrating high performance and robustness across ICU stays from 203 hospitals and 140,945 patients. These models have the capability to automate manual assessments while further allowing clinican staff to determine onset of ABD before it occurs. By doing so, patient outcomes in the ICU can be improved by reducing negative outcomes such as self-injury, coma, and death since clinican staff can intervene to prevent ABD before onset and in turn prevent negative outcomes caused by ABD.

# 1. Introduction

Acute brain dysfunction (ABD) in the intensive care unit (ICU), which includes delirium and coma, increases the chances of morbidity and mortality. It affects the majority (up to 64%) of critically ill patients at some point during their ICU stay [1-4] and is associated with higher mortality risk [5], longer hospital stays [6], and long-term cognitive impairment [7]. Current approaches for ABD diagnosis are limited to methods that assess brain status or conduct an overall risk assessment, which do not allow for preemptive and dynamic diagnosis of ABD in the ICU. Such approaches include the use of assessment scores such as the Glasgow Coma Scale (GCS) and Confusion Assessment Method (CAM), neuroimaging modalities such as magnetic resonance imaging (MRI), and biomarkers such as S100 calcium-binding protein B (S100B) and neuron-specific enolase (NSE) [8]. These approaches are typically administered at best only once during a 12-hour period. In contrast, a patient's condition in the ICU can deteriorate rapidly and fluctuate. It is essential to better capture the dynamic behavior of the patient's physiological state to assess the risk of ABD in the next hours and plan early interventions.

Given the wealth of data available in modern Electronic Health Record (EHR) systems, including vital signs, medications, laboratory results, assessment scores, and demographic data, data-driven approaches can predict ABD preemptively throughout a patient's stay. Deep learning techniques such as recurrent neural networks (RNNs) [9] and transformer models [10] have been shown to have excellent performance in dynamically predicting a variety of outcomes in the ICU, including organ failure, Sequential Organ Failure Assessment (SOFA) scores, readmissions, and mortality. These models can contextualize large amounts of data originating from EHR and patients' medical histories. This is essential in the case of predictive ICU models, as typical stays last from around three days to around a week, with many lasting 2-3 weeks. The recently developed selective state spaces model MAMBA [27] is even more suitable for such scenarios, with performance comparable to Transformer models, while still being able to process and learn from feature-rich datasets with less training time, linear scaling with sequence length, longer sequence lengths, and with a smaller model size.

Despite the importance and prevalence of ABD in the ICU, there are very few studies on the early prediction of this condition. Existing studies focus on predicting binary outcomes for delirium (*i.e.*, delirium or non-delirium). Several studies have attempted to predict the development of delirium at some point during the patient's stay in the ICU [11-14], postoperative delirium [15], and predicting delirium dynamically in the ICU [14]. To our knowledge, only two studies developed prediction models with multiple ABD outcomes. The first study focused on predicting next-day outcomes pertaining to coma, delirium, death in ICU, discharge, or normal brain function [3]. The second study used the same model as the first one to predict the transition from ABD (i.e., delirium and coma) to ABD-free states [4]. These studies also did not employ state-of-the-art deep learning models such as Transformers.
Additionally, these studies used small, private datasets originating from single cohort studies in individual hospitals, limiting their reproducibility and robustness. We trained and validated our models on a large dataset from the University of Florida (UF) Health and Shands Hospital to address these shortcomings. We externally validated its performance on two large, publicly available datasets, MIMIC-IV and eICU. This allowed us to demonstrate our models' generalizability between different hospital environments with

203 hospitals and 140,945 patients. Further, we evaluate our models on real-time prediction of ABD status in current patients with ongoing ICU stays at the UF Shands hospital as a prospective validation of their real-time performance.

## 2. Methods
### 2.1 Overview

To train our models, we used two public datasets (MIMIC, eICU) and the University of Florida Health (UFH) ABSTRACT ICU dataset (under IRB #201900354 and IRB #202101013). The ABSTRACT dataset is retrieved from the UFH Integrated Data Repository (IDR) and contains data captured from patients during routine critical care from 2014-2019. Further, we also analyzed a prospective dataset collected from patients in the UF Health Shands hospital whose ICU stays are ongoing to evaluate the real-time performance of our models in a practical ICU setting. All patients signed an informed consent form, or a legally authorized representative signed it on their behalf. Data was collected per the Declaration of Helsinki and relevant university rules and guidelines set by the UF Institutional Review Board (IRB).

All three datasets consist of vital signs (i.e., heart rate, blood pressure, respiratory rate, and ventilation), discrete codes (procedure, medication, laboratory, and diagnosis codes), and patient history. Patient history includes demographic information such as age, height, weight, sex, ethnicity, ongoing conditions, smoking, drinking habits, comorbidity, and other scores. Variables were divided into two types: static and dynamic. Each data type (static and temporal) is processed separately. Static data refers to all the features that do not change during an ICU visit, like patient history and demographics. Temporal data refers to variables whose values change during an ICU visit, such as vitals, medications, scores, and labs collected during the patient's ICU stay. Those medications and labs that represent less than 1% of the total unique medications and labs were removed from the dataset. The remaining temporal data is processed by standardizing their measurement units, imputing missing values using the median value during the ICU stay, and normalizing using the standard distribution. After preprocessing, static and temporal data are passed through separate embedding layers before input into the model. For models that cannot process temporal data, such as CatBoost, statistical features including mean, median, minimum, maximum, and standard deviation are calculated across each 12-hour interval of each ICU stay for all temporal features and then concatenated with the static data thus, the embedding layers are not used for these models.

Labels for each 12-hour interval are created using a brain phenotyping algorithm [26] developed utilizing the Richmond Agitation-Sedation Scale (RASS), Confusion Assessment Method (CAM), and the Glasgow Coma Scale (GCS) scores. These labels represent a patient's ABD status. Labels are also created for each transition between ABD states in a clinically relevant manner, as described below. The transition labels are divided into six categories- a) no-change (normal-normal, coma-coma, and delirium-delirium transitions), b) normal-delirium, c) normal-coma, d) delirium-coma, e) any abd-normal (i.e., delirium-normal and coma-normal), and f) any-death (a transition from either normal, coma, or delirium to death). By grouping transitions that share the same clinical meaning, we attempt to reduce the number of labels, increase the representation of each label, and eliminate the confusion that might arise from assigning different labels to similar physiological transitions.

Furthermore, combining all no-change labels into the same group will allow the model to learn physiological residuals over time instead of focusing on physiologic patterns associated with any one state. This is crucial as we train a separate model to classify the ABD states. Combining similar transitions reduces the number of labels, significantly reducing the confusion of model prediction between these states. Therefore, our models have two prediction heads, one for classifying the ABD state in the following

12-hour interval and the other for predicting the transition between the current 12-hour interval and the subsequent 12-hour interval. Figure 1 delineates an abstract representation of our method.

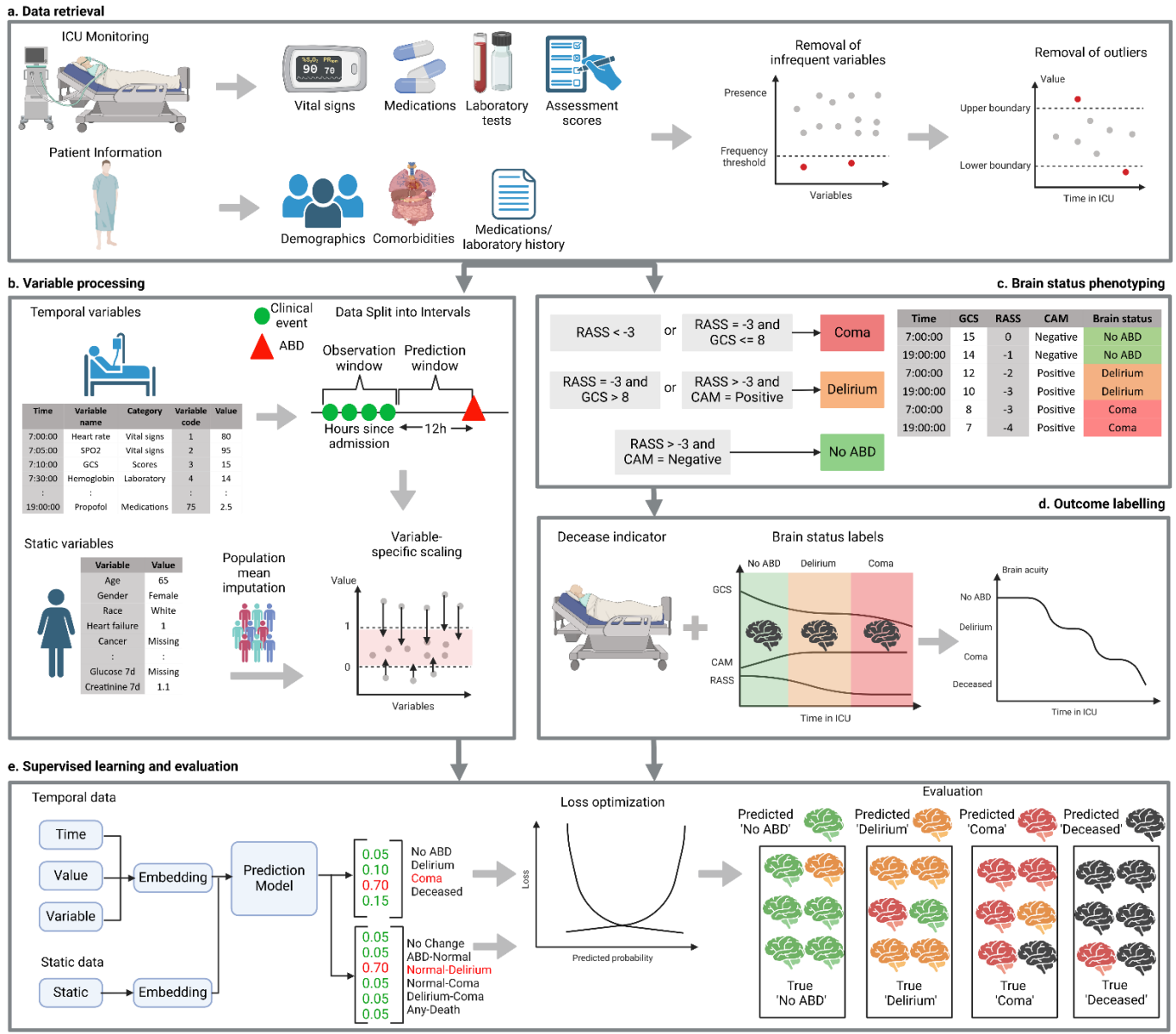

**Fig. 1: ABSTRACT model development overview.** (a). Data is retrieved from one private and two public intensive care unit (ICU) datasets (including vital signs, medications, laboratory test results, and assessment scores) and information/characteristics available before patient admission to the ICU (including patient demographics, comorbidities present at admission, and laboratory/medication history). Highly infrequent features and outliers for each feature are removed using an upper (99th percentile) and lower (1st percentile) bound. (b). Variables are processed according to their type. Temporal variables extracted from the ICU are merged into a table of clinical events, where each variable is assigned a code according to its order of appearance. A clinical event is composed of a triplet: the timestamp converted to hours since admission, variable code, and value. The events are split into intervals, where an observation window consists of all clinical events from admission up to the beginning of the prediction window, where ABD status will be assessed. The prediction window represents the interval within which the outcome is measured, with the window length being 12 hours. After splitting, each variable is scaled using standard distribution scaling. Static variables extracted from patient information are first imputed using the population mean for each feature and subsequently scaled using the standard distribution scaling. (c). The ABD phenotypes are calculated every 12 hours in the ICU using the definitions of GCS, RASS, and CAM. (d). The outcome for each 12-hour window is determined by combining ABD status labels with deceased indicators, which provide a complete trajectory of brain acuity for each patient. These labels are then used to create a separate label for the transition between ABD states. (e). A transformer-based and MAMBA model is employed for supervised learning of brain acuity prediction, where two embeddings are created for temporal and static data. The first embedding goes through a transformer encoder to compute attention between clinical events in a sequence, from which the output is then combined with the second embedding to calculate the probabilities of each outcome. The loss of the model is optimized by comparing the predicted probabilities with the ground truth labels. Finally, the model is evaluated by calculating the classification error for each outcome and transition individually.

## 2.2 Data Curation

Data curation is kept the same across the different datasets to ensure accurate comparisons. A specific ICU stay is only included in the study if it is at least 24 hours and contains at least one measurement for RASS, GCS, or CAM scores. The ICU stays, which resulted in mortality but did not contain the required scores, are not included in our analysis. For the sake of consistency, homologous features from different datasets are manually mapped to each other and collected in the same way to ensure uniformity in the data processing.

The ABSTRACT dataset collected at UF Health Shands hospital contains a more extensive patient history including information about past medication, labs, smoking status, insurance type, and admission source (transfer from another hospital or emergency room), which are unavailable for eICU [25] and MIMIC-IV [24] datasets. Therefore, these features which were not omnipresent in all datasets were excluded from the analysis to maintain consistency. Table 1 shows the features collected for processing from each dataset.

## 2.3 Data Processing
### 2.3.1 ABD Phenotype

ABD status is determined for each 12-hour interval using phenotyping logic previously developed [26]. The ABD phenotype is determined using logical rules based on the RASS, CAM, and GCS scores measured during each 12-hour interval, shown in supplemental Figure 1H. Each score is imputed using the forward fill method from the previous 12-hour interval only if another score was measured around the same time. For example, if for one 12-hour interval, there is a recorded value for RASS, GCS, and CAM, but the next 12-hour interval has records for RASS and GCS but not CAM (measured around the same time of day as the previous interval), the previous CAM score is forward-filled to the next 12-hour interval. Missing values for consecutive 12-hour intervals are not imputed; meaning values are not forward-filled across multiple 12-hour intervals, only the immediate interval after a recorded score value. Mortality labels are added separately to obtain the final label for the outcome of each 12-hour interval.

### 2.3.2 State Transition Labels

Additional labels are created using the outcome label for each 12-hour interval, signifying the transition between two consecutive intervals. This is necessary because of the high probability for patients to stay in an ABD state between consecutive 12-hour intervals. Transitions are also when clinical intervention is needed, which is essential for a real-time prediction system to predict these transitions accurately. The coma-delirium transition is excluded from these transition labels since patients who are already in a coma are expected to require more intensive care when the patient comes out of the comatose state; as such, it is not as relevant to clinicians as the other transitions between states. Including these labels with outcome prediction improves model performance for the most critical transitions.

### 2.4 Modeling and Analysis

The three datasets that are used in this study- a) eICU (contained data from ~200 hospitals), b) MIMIC (data collected from Beth Israel Deaconess Medical Center; Boston, MA), and c) ABSTRACT (data collected from UF Health Shands hospital) are processed in a manner described in the preceding paragraphs. Then, these datasets were randomly divided into a development cohort of 80% of the total number of patients with all their ICU stays and a test cohort of the remaining 20% of the patients. Five-fold cross-validation is performed on the development cohort to find the best deep-learning model parameters. These models were then trained on the entire development set and evaluated on the holdout test set, calculated using a 95% confidence interval across the five different fold's performance on the holdout test set. For comparing performance across datasets, models in this study were trained using a dataset from one source, such as eICU, and then evaluated on another dataset source, such as UF Shands. The cost function for training these models was modified to penalize false positives more such that they incur larger error and, consequently, larger parameter updates when false positives occur to reduce the prevalence of early detections of transitions between ABD states.

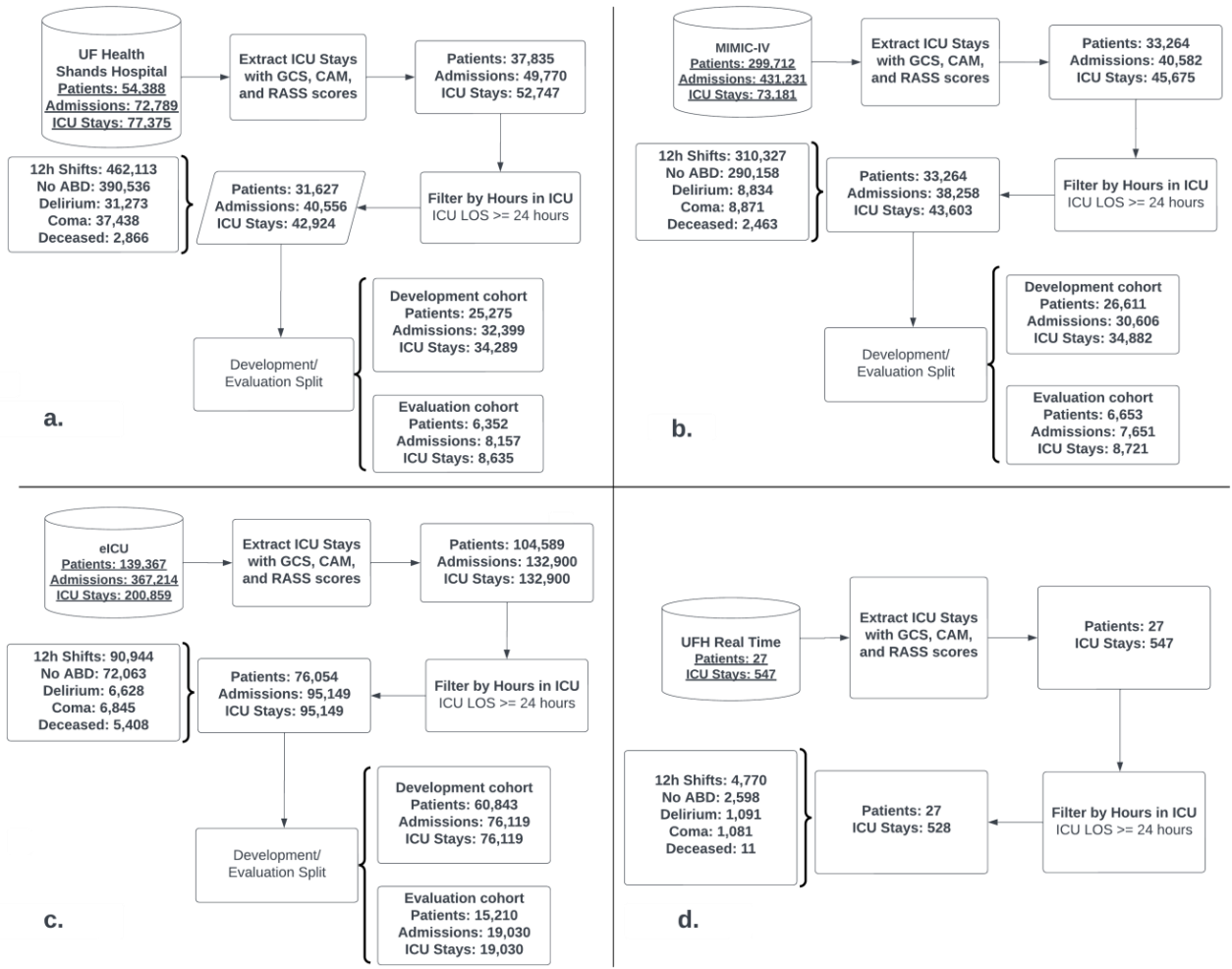

**Fig. 2: Datasets; (a) UFH ABSTRACT, (b) MIMIC-IV, (c) eICU, and (d) UFH ABSTRACT Real Time.** The development cohort is used for 5-fold cross-validation training for each model, where it is split using 80% training and 20% validation sets. After training on each fold, the evaluation cohort is used to evaluate and report model performance.

### 2.4.1 Baseline Models

Two baseline models are used to compare the performance with the proposed models: the CatBoost [22] and the Gated Recurrent Unit (GRU) network [23]. The CatBoost model [22], which is built on an ensemble approach using gradient-boosted decision trees, is notable for its high performance with minimal need for fine-tuning of hyperparameters. This model is designed to work with statistical features (mean, median, minimum, maximum, and standard deviation) that summarize each 12-hour period. To incorporate medication and laboratory data, we transform these into distinct features for each medication and lab test. These are then one-hot encoded to indicate the presence of each medication and lab test within a given 12-hour period. The dataset for the CatBoost model is organized into a data frame, with each row corresponding to a 12-hour interval. The columns of this dataframe include vital signs, scores, medications, lab tests, static features, and a timestamp marking the start of the 12-hour interval.

The second baseline model is a GRU model containing 4 GRU layers of 200, 200, 100, and 50 GRU units with a 20% dropout after each layer and two output heads for predicting outcome and transition for each

12-hour interval. GRU is a class of RNNs that provides a baseline to evaluate the performance improvement when using a Transformer architecture over RNNs. GRU data processing uses a limited sequence length for each 12-hour interval of 512, while for the Longformer model, sequence lengths can vary for each interval, up to 4096. The longest sequence length in the datasets used is 1200.

### 2.4.2 Transformer Model

The Longformer [21] model is a modified transformer model that incorporates the use of local and global attention mechanisms to achieve linear scaling of attention complexity with sequence length, versus the quadratic scaling of attention complexity found in the vanilla transformer architecture. This is useful for processing EHR data which tends to have longer sequence lengths and contains many features. The max sequence length for the Longformer architecture is 4096, while the max sequence length in our datasets is 1200. The Longformer architecture including embedding layers is shown in supplemental Figure 3H.

### 2.4.3 MAMBA Model

The MAMBA model [27] is a state-of-the-art selective state space model that performs similarly to transformers that can scale linearly with sequence length. The model utilizes a structured state space model (SSM) where the SSM parameters are parameterized based on the input, allowing for the model to remember relevant information indefinitely. This change creates a technical challenge when performing computation of the model, as all prior SSM models had to be time and input-invariant. The authors created a hardware-aware computation algorithm to efficiently compute the model and scale linearly with sequence length. The architecture of each MAMBA block combines the MLP block of transformer models with the SSM architecture, creating a fully recurrent model. Our proposed MAMBA model uses the same embedding layers and data preprocessing as the Longformer model, as shown in Figure 3.

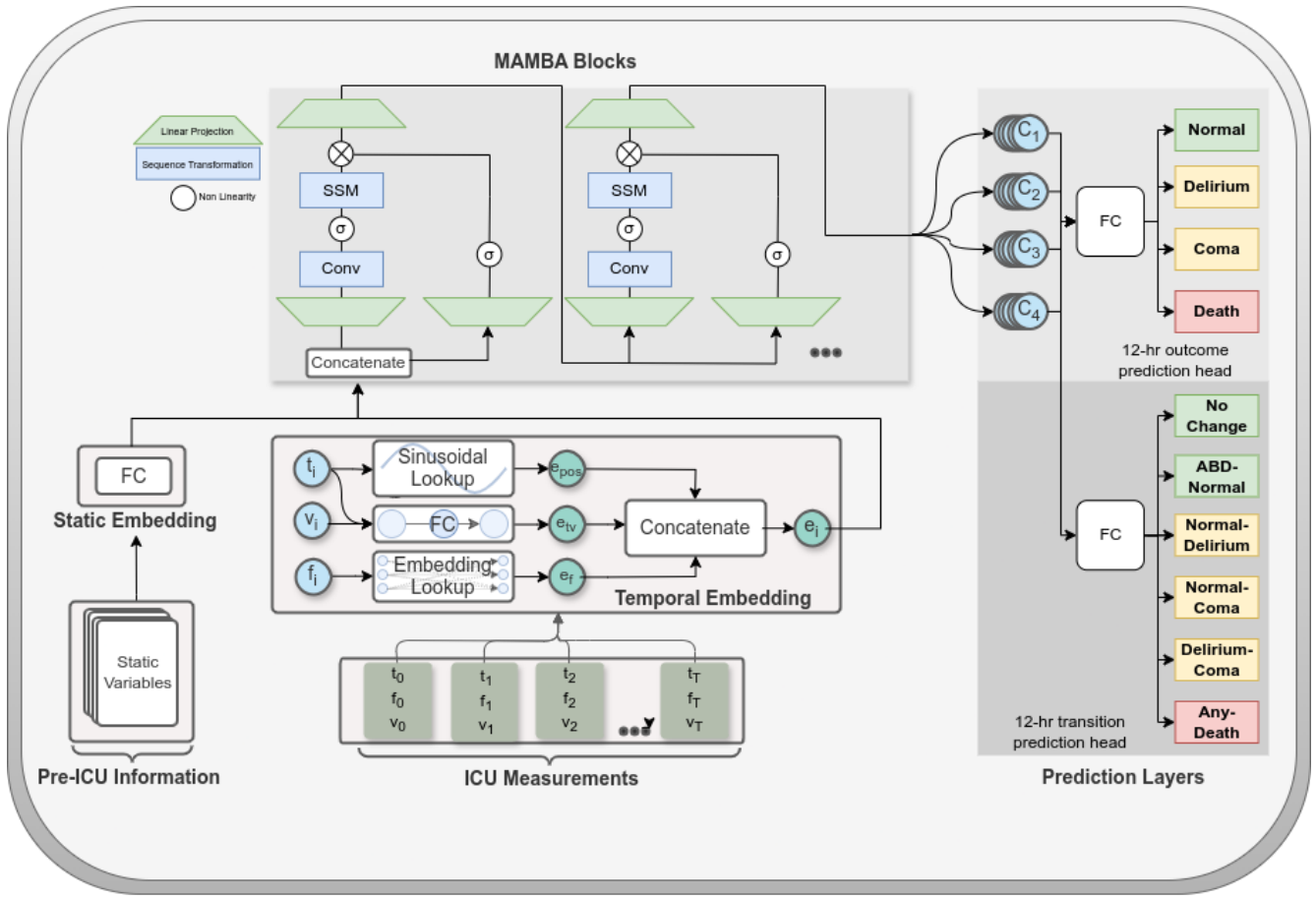

**Fig 3: MAMBA model architecture for predicting ABD status in 12-hour windows.** Pre-ICU information includes a summary of patient medications and laboratory tests before admission and sociodemographic indicators. The Temporal ICU measurements take a flexible, variable form list of tuples, each containing the measurement time (t), measurement value (v), and a variable identifier(f) that specifies the type of data the tuple represents. Static variables are processed through an embedding FC layer and concatenated with the temporal embedding. Output tokens are concatenated and passed through an FC layer following an output layer that predicts ABD outcome and a separate FC layer and outcome layer for predicting ABD transition, both using a softmax activation.

## 3. Results

### 3.1 Participants

The three datasets used in this study had similar patient demographics from distinct hospitals; over 200 for eICU, UF Health Shands, and Beth Israel Deaconess Medical Center (Boston, MA). To properly validate model performance between the datasets, comparisons in Figure 2 and Table 1 describe each dataset used in this study and their respective cohorts. Across each dataset, ABD affects older patients who typically have some precondition, shown in Table 1 with comorbidities scores.

**Table 1: Cohort statistics for UF, MIMIC, and eICU.** General statistics represent the entire dataset after filtering LOS. Statistics include patients who experience ABD at least once during an ICU stay.

| Cohort Characteristics | ABSTRACT | | MIMIC-IV | | eICU | |
|---|---|---|---|---|---|---|
| | All | ABD | All | ABD | All | ABD |
| Total patients, n | 31,627 | 9,347 | 33,264 | 11,024 | 76,054 | 6,294 |
| Total ICU stays, n | 42,924 | 10,416 | 43,603 | 12,746 | 95,149 | 6,935 |
| Age, mean (std) | 59.6 (17.0) | 60.8 (16.2) | 64.3 (16.5) | 64.8 (16.2) | 62.7 (16.2) | 64.3 (15.4) |
| Gender, Male, n (%) | 17,672 (55.9%) | 5,413 (57.9%) | 19,039 (57.2%) | 6,317 (57.3%) | 41,176 (54.1%) | 3,449 (54.8%) |
| Gender, Female, n (%) | 13,955 (44.1%) | 3,934 (42.1%) | 14,225 (42.8%) | 4,707 (42,7%) | 34,858 (45.8%) | 2,845 (45.2%) |
| Race, White, n (%) | 24,043 (76.0%) | 6,899 (73.8%) | 22,511 (67.7%) | 6,875 (62.4%) | 58,611 (77.1%) | 4,926 (78.3%) |
| Mean BMI (std) | 28.5 (7.7) | 28.4 (7.8) | 16.1 (15.2) | 18.2 (15.0) | 29.3 (8.6) | 29.1 (8.7) |
| ICU length of stay, mean | 5 days 7 hours | 11 days 13 hours | 4 days 14 hours | 7 days 10 hours | 3 days 22 hours | 6 days 6 hours |
| Charlson comorbidity, mean (std) | 2.4 (2.6) | 2.8 (2.7) | 0.2 (1.1) | 0.4 (1.4) | 3.9 (2.8) | 4.2 (2.8) |
| Cerebrovascular disease, n (%) | 4,890 (15.5%) | 1,987 (21.3%) | 328 (1.0%) | 218 (2.0%) | 8,277 (8.7%) | 762 (11.0%) |
| Congestive heart failure, n (%) | 7,937 (25.1%) | 2,695 (28.8%) | 653 (2.0%) | 372 (3.4%) | 14,926 (15.7%) | 1180 (17.0%) |
| Dementia, n (%) | 1,211 (3.8%) | 571 (6.1%) | 10 (0.01%) | 9 (0.1%) | 3,100 (3.3%) | 409 (5.9%) |
| Mild liver disease, n (%) | 2,701 (8.5%) | 1,166 (12.5%) | 426 (1.3%) | 280 (2.5%) | 1,862 (2.0%) | 174 (2.5%) |
| Renal disease, n (%) | 6,124 (19.4%) | 2,247 (24.0%) | 1,010 (3.0%) | 540 (4.9%) | 11,994 (12.6%) | 953 (13.7%) |
| Total patients deceased, n (%) | 2,865 (9.1%) | 2,247 (24.0%) | 2,089 (6.3%) | 1,495 (13,6%) | 5,408 (5.7%) | 681 (9.8%) |

## 3.2 Outcome Prediction

The models performed exceptionally well in predicting the outcome for each 12-hour interval, with the Longformer and MAMBA models performing similarly. Performance is comparable to current literature, although comparisons have limitations since our models dynamically predict ABD in 12-hour intervals. Model performance on the ABSTRACT dataset is shown in Table 2 for predicting the outcome of the following 12-hour interval given processed EHR data of the patient during their ICU stay before the 12-hour interval occurs. Table 3 then shows model performance when evaluated against eICU and MIMIC-IV datasets and vice versa. Supplemental Tables 1H and 2H show the performance of our proposed models against baseline architectures. The first prediction occurs at least 12 hours after admission. The outcome for a 12-hour interval is determined using the phenotyping logic and collected scores during the entire 12-hour interval. These scores are typically recorded around the beginning of a nurse's 12-hour shift (7 a.m. and 7 p.m., respectively), although only some scores are recorded around that time. All scores recorded during each 12-hour interval are used to determine the patient's ABD status.

Further, we compared model performance among different demographics, including age, gender, and ethnicity. Results are averaged across five different folds, and our models perform better than the current literature for delirium prediction [4, 15, 17], although these metrics are not for dynamic prediction. To evaluate model performance throughout an ICU stay, we calculate AUC throughout each 12-hour interval averaged across all ICU stays in each dataset for predicting outcomes.

### 3.3 Transition Prediction

To show that the model can predict critical transitions between ABD states, rather than just the outcome of the following 12-hour interval's state (which transitions between states are seldom), the models were trained and evaluated using transition labels representing the most critical transitions between 12-hour intervals. Table 2 shows results on the ABSTRACT dataset, with Table 3 showing the evaluation of the eICU and MIMIC-IV datasets. Transition prediction throughout an ICU stay is shown in Figure 5, generated using the Longformer model on each dataset. These results represent model performance in a real-time setting, where predicting a transition between ABD states can allow for intervention before the onset of ABD.

**Table 2:** ABD prediction results from 5-fold cross-validation on ABSTRACT dataset, expressed as the Area Under the Receiving Operator Characteristic Curve (AUROC). Each value has a 95% confidence interval (CI) calculated from the test set's performance for each fold. The testing set is a separate set, detailed in Figure 2, taken before cross-validation is applied. Trained models are optimized to the validation set of each fold, not the separate test set.

|  | **Longformer** | **MAMBA** |
|---|---|---|
| Outcome (12-hour window) |  |  |
| Coma | 0.98 (0.98-0.99) | **0.99 (0.98-0.99)** |
| Delirium | 0.90 (0.90-0.90) | **0.91 (0.91-0.91)** |
| Deceased | **0.99 (0.90-0.90)** | 0.90 (0.90-0.90) |
| Transition (12-hour window) |  |  |
| No Change | **0.80 (0.79-0.80)** | 0.78 (0.78-0.78) |
| ABD-Normal | **0.85 (0.84-0.85)** | 0.84 (0.84-0.84) |
| Normal-Delirium | 0.71 (0.71-0.71) | **0.73 (0.72-0.73)** |
| Normal-Coma | 0.79 (0.79-0.79) | **0.81 (0.81-0.82)** |
| Delirium-Coma | 0.78 (0.78-0.78) | 0.78 (0.78-0.78) |
| Any state-Death | **0.85 (0.85-0.85)** | 0.84 (0.84-0.84) |

**Table 3:** External validation of models on ABSTRACT, eICU, and MIMIC data sources for the **MAMBA** model. The method denotes the dataset used for training and the dataset used for evaluation. Training is done using 5-fold cross-validation on the specified dataset. For example, ABSTRACT training and ABSTRACT testing denotes training and testing done using just the UF-ABSTRACT Shands dataset, while ABSTRACT training and eICU testing denote training using the ABSTRACT dataset and evaluating performance on the entirety of the eICU compiled dataset.

|  | **ABSTRACT** | **MIMIC-IV** | **eICU** | **ABSTRACT/eICU** |
|---|---|---|---|---|
| Outcome (12-hour window) |  |  |  |  |
| Coma | 0.99 (0.98-0.99) | 0.95 (0.94-0.95) | 0.97 (0.97-0.97) | 0.91 (0.90-0.92) |
| Delirium | 0.91 (0.91-0.91) | 0.89 (0.89-0.89) | 0.92 (0.92-0.92) | 0.88 (0.88-0.89) |
| Deceased | 0.90 (0.90-0.90) | 0.92 (0.91-0.92) | 0.91 (0.91-0.91) | 0.89 (0.89-0.90) |
| Transition (12-hour window) | | | | |
| No Change | 0.78 (0.78-0.78) | 0.76 (0.75-0.76) | 0.79 (0.79-0.79) | 0.74 (0.74-0.75) |
| ABD-Normal | 0.84 (0.84-0.84) | 0.86 (0.86-0.86) | 0.88 (0.87-0.88) | 0.81 (0.81-0.81) |
| Normal-Delirium | 0.73 (0.72-0.73) | 0.70 (0.70-0.70) | 0.75 (0.74-0.75) | 0.69 (0.69-0.70) |
| Normal-Coma | 0.81 (0.81-0.82) | 0.80 (0.80-0.80) | 0.81 (0.81-0.81) | 0.86 (0.86-0.86) |
| Delirium-Coma | 0.78 (0.78-0.78) | 0.75 (0.75-0.76) | 0.77 (0.77-0.77) | 0.74 (0.74-0.75) |
| Any state-Death | 0.84 (0.84-0.84) | 0.82 (0.82-0.82) | 0.84 (0.84-0.85) | 0.82 (0.82-0.83) |

### 3.3.1 False positives due to Early Predictions

When analyzing confusion matrices and false positive rates for transition predictions, we noticed high confusion when the model predicts a transition between a normal and an ABD state as a false positive. Further investigation revealed that the models tended to predict the transition from normal to an ABD state in 2-4 12-hour intervals before the onset of ABD occurs, which is more prevalent in the Longformer model than the MAMBA model. This issue is shown in the confusion matrix depicted in Figure 7. To account for this, we split transitions into "episodes," where the time between transitions to and from ABD states is determined. We then recalculate the metrics by determining the model to be "correct" when it correctly predicts the start or end of an episode up to 48 hours before the transition occurs. By doing so, model performance improves significantly, from 0.73 to 0.79 AUROC for the normal-delirium transition and 0.81 to 0.85 for the normal-coma transition, as shown in Table 4. Figure 7 further shows that false positive rates increase when predicting ABD transitions as model predictions approach any 12-hour interval.

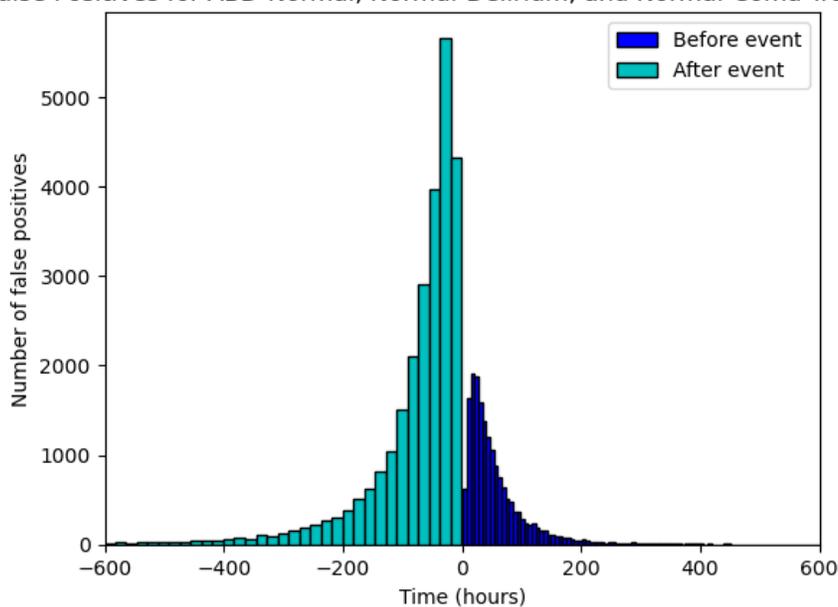

**Fig. 4: False positive graph showing false positive predictions of transitions for normal-delirium and normal-coma transitions, evaluated on the UFH ABSTRACT dataset and the MAMBA model.** There is a high false positive rate immediately before and after the transition event, which mainly falls

within 48 hours of the event. This demonstrates that the model tends to predict the onset of ABD much earlier than it occurs.

**Table 4:** Metrics calculated with a 48-hour lead time for model predictions. Model predictions that occur within 48 hours of an event are counted as correct predictions. Model performance improves substantially with a 48-hour lead window, suggesting that the MAMBA model predicts the onset of ABD earlier than the 12-hour interval where ABD onset occurs.

|  | MAMBA | MAMBA 48-hr Lead |
|---|---|---|
| Outcome (12-hour window) |  |  |
| Coma | 0.99 (0.98-0.99) | 0.99 (0.99-0.99) |
| Delirium | 0.91 (0.91-0.91) | **0.93 (0.93-0.93)** |
| Deceased | 0.90 (0.90-0.90) | **0.93 (0.93-0.93)** |
| Transition (12-hour window) |  |  |
| No Change | 0.78 (0.78-0.78) | **0.81 (0.81-0.81)** |
| ABD-Normal | 0.84 (0.84-0.84) | **0.85 (0.85-0.85)** |
| Normal-Delirium | 0.73 (0.72-0.73) | **0.79 (0.79-0.80)** |
| Normal-Coma | 0.81 (0.81-0.82) | **0.85 (0.85-0.85)** |
| Delirium-Coma | 0.78 (0.78-0.78) | **0.81 (0.81-0.81)** |
| Any state-Death | 0.84 (0.84-0.84) | **0.89 (0.88-0.89)** |

### 3.4 Real Time Deployment

To further verify the model's performance in a simulated real-time setting, we obtained data from current patients enrolled in our study from the UFH IDR, containing data from May to July 2023. Table 5 shows the prediction performance for this prospective dataset. Refer to Figure 2 for statistics on this dataset.

**Table 5:** ABD prediction results on the prospective real-time dataset, showing mean AUROC for all classes and individual AUROC metrics for each outcome and transition.

|  | Longformer | MAMBA |
|---|---|---|
| Outcome (12-hour window) |  |  |
| Coma | 0.98 | **0.99** |
| Delirium | 0.89 | **0.90** |
| Deceased | 0.99 | 0.97 |
| Transition (12-hour window) |  |  |
| No Change | **0.87** | 0.86 |
| ABD-Normal | 0.84 | **0.91** |
| Normal-Delirium | 0.77 | **0.84** |
| Normal-Coma | 0.81 | **0.83** |
| Delirium-Coma | **0.72** | 0.69 |
| Any state-Death | **0.89** | 0.88 |

## 4. Discussion

By employing multiple datasets from various sources, including evaluation using one dataset to train and another for testing, we show that our models can reliably predict the onset of ABD during a typical stay in an ICU. Through this research, we discovered that predicting the outcome and precisely predicting critical transitions between ICU states is crucial. When we evaluated how well the model predicts a transition between ABD states without transition labels, we saw that the models' performance could have been better, with only a 0.634 mean AUROC on the ABSTRACT dataset. We attribute this to the nature of patients in an ABD state tending to stay in that stay for consecutive 12-hour intervals, biasing the model to predict the current state of ABD the patient is in for every consecutive 12-hour interval, reducing performance. By creating the transition labels and adding another prediction head to each model, outcome prediction performance increased, and the prediction of critical ABD transitions considerably increased. These critical transitions between ABD states are crucial for our models to predict when deployed in a real-time setting to prevent adverse outcomes resulting from ABD.

Further, there are limitations with the phenotyping we used to determine ABD status for patients. While RASS, CAM, and GCS scores are significant indicators of a patient's mental status, they are biased by those administering the test and the patient's pre-existing conditions. For example, suppose a patient has a baseline of reduced mental status. In that case, their baseline scores for CAM and GCS will reflect this and negatively bias our model with features indicating a normal ABD status but with corresponding CAM and GCS scores signifying reduced mental state; however, we cannot determine how often this may occur in each dataset. Evaluation between datasets also shows there may be differences in how these tests are administered by nurses in the ICU, which can affect model performance negatively. Improving the phenotyping logic using patient history and possibly other mental status scores, such as Sequential Organ Failure Assessment (SOFA) and Modified Early Warning Score (MEWS), can improve the labeling of ABD and model performance. Future studies should explore this, alongside focusing on predicting transition between ABD states rather than outcomes, as we have demonstrated it is not enough to predict only the outcome for a given time interval when using a model meant to be deployed for real-time use.

These limitations may also account for the high false positive rate in predicting the transitions between ABD states, specifically with models that predict 24-48 hours before the transition between normal and an ABD state. We also investigated using shorter and longer prediction intervals during training, from 3 hours to 48 hours, with the performance of all models increasing as the prediction window became smaller. Because of this, for our task of being able to deploy these models for real-time prediction of the onset of ABD, we can consider model performance to be much better than what the results show, as the model's task of an early warning system for ABD is accomplished even if the model predicts the onset of ABD much earlier than 12 hours. In other studies we are currently working on, we noticed a similar issue with predicting patient acuity during their ICU stay, where models tend to predict acuity 24-48 hours before onset when trained to predict on 12-hour intervals.

Our study investigates using state-of-the-art Transformers and structured state-space architectures over other models, which have yet to be widely researched for the prediction of delirium, coma, or general ABD in current studies. We propose that future studies use the latest transformer or comparable models in their investigations to improve performance on these tasks and allow for larger datasets with more features to be evaluated. Future work should explore more testing performance of models deployed in a real-time setting as we speculate they may perform better than current results based on large numbers of early predictions. Predicting the onset of ABD even earlier than 12 hours, which our false positive rates show, is still clinically relevant regardless of performance metrics. Our work provides a solid basis for more exploration into the dynamic prediction of ABD.

## 5. Contributions

Our methods contribute five improvements over existing methods: (1) using a tighter prediction window of 12 hours compared to 24 hours as in literature to allow for quicker intervention time, (2) using more complex models over simpler models that allow for processing of more features and extended sequences of temporal data, (3) predicting the ABD phenotype versus the single delirium outcome as in literature, (4) validating our methods across three large datasets to show model robustness and transferability between ICU wards of different hospitals, and (5) predicting transition between ABD states as well as outcomes for every 12-hour interval.

A window of 24 hours or more is used in various studies [17, 19], especially with PRE-DELIRIC and its derivations, which can miss delirium onset before this window, leading to adverse outcomes associated with delirium and comatose. These models are designed for something other than dynamic delirium prediction as well, only using a sample of data to predict delirium onset on the entire ICU stay for a patient, where our methods can predict delirium in 12-hour dynamic windows.

We evaluated transformer-based approaches for this task, performing better than traditional machine learning methods like CatBoost, Random Forests, and XGBoost. These statistical machine learning methods are smaller and faster to train; however, they cannot represent temporal relationships among the data points, which Transformer models excel at. Most models evaluated in other studies are these ensemble models, logistic regression, or small deep-learning models [17-20]. As shown in our study, Transformer models can outperform these other methods.

While outperforming similar studies, our model also predicts related outcomes besides delirium, comatose, and mortality, in which very few studies use these same outcomes. Another study was found, which showed comparable results to ours [3]. Our methods outperform delirium prediction and allow for dynamic prediction of coma and mortality.

With this, all of our models are evaluated on a large dataset obtained at UF Shands, as well as two publicly available datasets, MIMIC-IV and eICU. Most other studies use small, private datasets [16], which makes their methods harder to recreate. By validating against multiple datasets from over 200 hospitals, we show our models that can perform in real-time without requiring training on data from a particular hospital. Further, we show that models can perform similarly to other data from different hospitals not used in training. This shows that our study is reproducible, demonstrating model robustness and generalizability using publicly available data.

We noticed high outcome prediction performance while evaluating model performance during this study. However, when examining the transition between ABD states, outcome performance could have been better, as the model tends to predict the current ABD state of each patient for every subsequent 12-hour interval. This is heavily influenced by the low number of transitions in each dataset, as patients tend to stay in their ABD state much more often than the transition between them, which is the most critical to capture for a real-time prediction model. By adding another prediction head to each model to predict clinically relevant transitions, we not only improve the model performance slightly on outcome prediction, but we also see significant improvements in predicting transition between ABD states, something current literature fails to address. Coupled with our real-time prospective results, we demonstrate the applicability of our models to alert clinicians about ABD onset for patients well before onset occurs.

## 6. Conclusion

Our study addresses a critical need for dynamic early prediction of acute brain dysfunction in the ICU to improve patient outcomes and reduce the cost of care. ABD, including delirium and coma, poses significant challenges for clinicians, given its association with increased patient mortality, extended hospital stays, and long-term cognitive impairment. Current manual diagnostic methods, which rely on infrequent testing done by nurses, are not robust enough to catch ABD early enough for intervention. To fill this need, our study leverages the vast amount of data available as electronic health records, processing vitals, medications, laboratory results, assessment, scores, and patient demographics to predict ABD before onset starts, demonstrating the potential to predict ABD in the ICU dynamically and offering a proactive approach for early intervention. Our study does have limitations that hinder performance, such as the phenotyping logic used to determine the ABD status of patients. ABD can be influenced by many factors that the phenotype does not address, which can reduce the accuracy of the phenotype and increase bias in the models when training on noisy labels. Future research should investigate different phenotyping logic for ABD and perform more comprehensive testing in a real-time setting to evaluate the model's performance empirically in a practical, real-world setting. Overall, we provide a foundation for future research to investigate ABD in ICU stays, leveraging the large number of EHRs available to train powerful deep-learning models to predict ABD onset with enough lead time for early intervention.

## 8. Data Availability Statement

## 9. Acknowledgements

The authors were supported by the National Institutes of Health (NIH) to conduct this study under award number 5R01NS120924-02 and 5R01EB029699-02.

## 10. Supplemental Figures

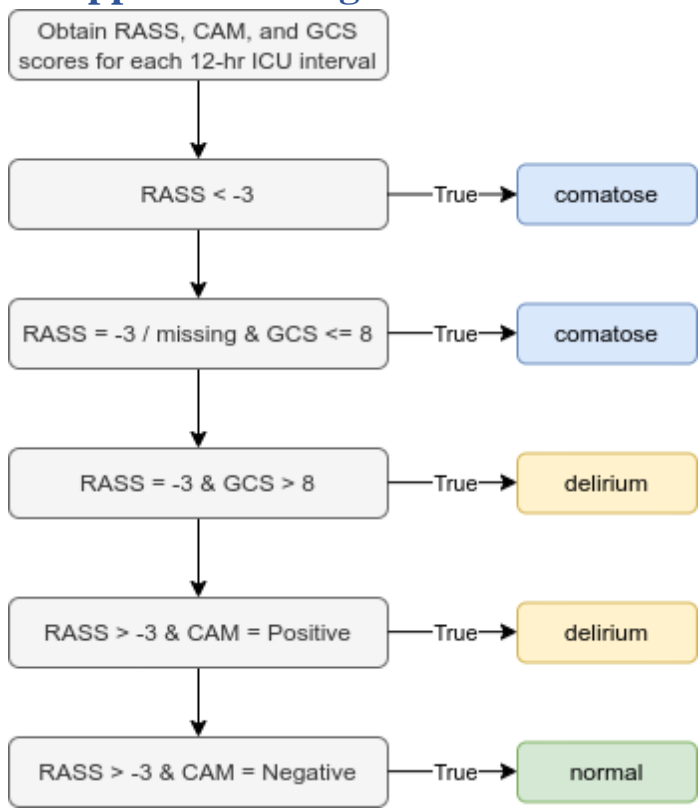

**Fig. 1H: Logic flow for generating ABD phenotype labels.** The phenotyping logic uses RASS, CAM-ICU, and GCS scores to determine ABD for each patient. The first rule is checking the RASS score for comatose, which is signified by a RASS score of less than –3. If RASS is missing or equals –3, the GCS score is also used to determine comatose status, with a score less than or equal to 8 signifying comatose. If GCS is greater than 8 with a RASS score of –3, then the patient is determined to be in a delirium state. Finally, if the RASS score is greater than –3 or missing, then the delirium state is specified using the CAM-ICU Score, with a positive CAM-ICU score stating that the patient is in delirium. In contrast, a negative score indicates a normal state.

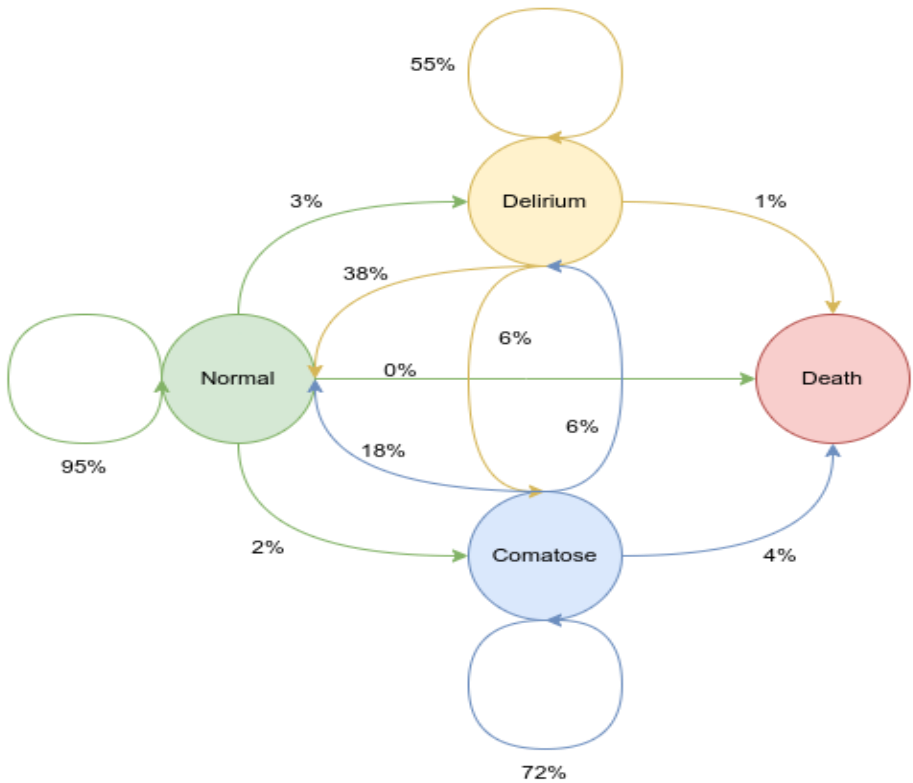

**Fig. 2H: State transition diagram for each possible ABD state.** The normal state signifies that the patient is not in any other ABD state. The death state is a final state where no other transitions from this state will occur. Probabilities are calculated using state transitions between consecutive 12-hour windows. Patients suffering from ABD tend to stay in their state more often between each 12-hour interval, as shown by high probabilities for staying in each ABD state.

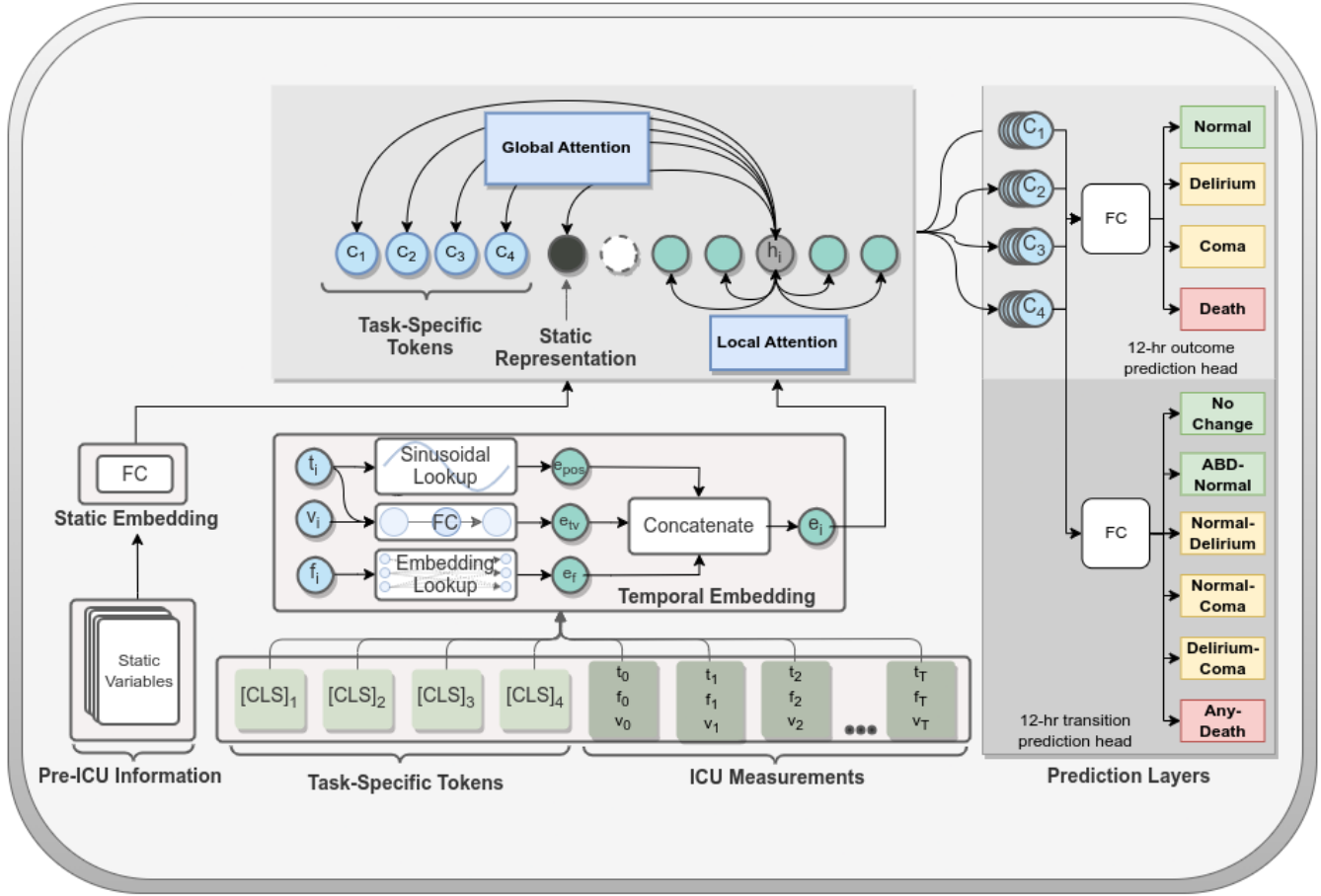

**Fig. 3H: Longformer model architecture for predicting ABD status in 12-hour windows.** Pre-ICU information includes a summary of patient medications and laboratory tests before admission and sociodemographic indicators. The Temporal ICU measurements take a flexible, variable form list of tuples, each containing the measurement time (t), measurement value (v), and a variable identifier(f) that specifies the type of data the tuple represents. Task-specific [CLS] tokens are assigned a value for time of prediction and are embedded before passing through a stack of Longformer layers with sliding self-attention windows. Global attention is then applied to both the predictor tokens and static features. Tokens are concatenated and passed through an FC layer following an output layer that predicts ABD outcome and a separate FC layer and outcome layer for predicting ABD transition, both using a softmax activation.

**Table 1H: Baseline model CatBoost performance on each dataset.** Performances shown for internal and external validation datasets for the CatBoost model.

|  | ABSTRACT | MIMIC-IV | eICU | ABSTRACT/eICU |
|---|---|---|---|---|
| Outcome (12-hour window) | | | | |
|    Coma | 0.95 (0.94-0.95) | 0.91 (0.91-0.91) | 0.94 (0.93-0.95) | 0.92 (0.91-0.93) |
|    Delirium | 0.81 (0.80-0.82) | 0.74 (0.73-0.74) | 0.78 (0.78-0.78) | 0.71 (0.70-0.72) |
|    Deceased | 0.97 (0.97-0.98) | 0.92 (0.92-0.92) | 0.96 (0.96-0.96) | 0.95 (0.94-0.95) |
| Transition (12-hour window) | | | | |
| No Change | 0.71 (0.71-0.71) | 0.69 (0.69-0.69) | 0.72 (0.71-0.72) | 0.70 (0.69-0.70) |
| ABD-Normal | 0.72 (0.72-0.73) | 0.74 (0.74-0.74) | 0.70 (0.70-0.70) | 0.67 (0.66-0.67) |
| Normal-Delirium | 0.62 (0.61-0.63) | 0.63 (0.62-0.63) | 0.62 (0.61-0.62) | 0.59 (0.59-0.60) |
| Normal-Coma | 0.65 (0.65-0.66) | 0.61 (0.60-0.62) | 0.64 (0.63-0.65) | 0.61 (0.61-0.62) |
| Delirium-Coma | 0.68 (0.68-0.68) | 0.66 (0.66-0.66) | 0.66 (0.65-0.66) | 0.58 (0.58-0.58) |
| Any state-Death | 0.79 (0.79-0.79) | 0.78 (0.78-0.78) | 0.79 (0.79-0.79) | 0.75 (0.75-0.75) |

**Table 2H: Baseline model GRU performance on each dataset.** Performances shown for internal and external validation datasets for the GRU model.

|  | **ABSTRACT** | **MIMIC-IV** | **eICU** | **ABSTRACT/eICU** |
|---|---|---|---|---|
| Outcome (12-hour window) | | | | |
| Coma | 0.96 (0.95-0.96) | 0.96 (0.96-0.96) | 0.97 (0.97-0.97) | 0.92 (0.91-0.92) |
| Delirium | 0.79 (0.76-0.82) | 0.80 (0.79-0.80) | 0.80 (0.80-0.80) | 0.74 (0.74-0.75) |
| Deceased | 0.95 (0.94-0.96) | 0.94 (0.94-0.95) | 0.95 (0.95-0.96) | 0.90 (0.90-0.91) |
| Transition (12-hour window) | | | | |
| No Change | 0.74 (0.74-0.74) | 0.72 (0.72-0.72) | 0.72 (0.72-0.73) | 0.69 (0.69-0.70) |
| ABD-Normal | 0.76 (0.75-0.77) | 0.78 (0.78-0.78) | 0.76 (0.76-0.76) | 0.72 (0.72-0.73) |
| Normal-Delirium | 0.66 (0.65-0.66) | 0.64 (0.64-0.65) | 0.68 (0.67-0.68) | 0.64 (0.64-0.65) |
| Normal-Coma | 0.71 (0.71-0.72) | 0.71 (0.71-0.72) | 0.72 (0.72-0.72) | 0.66 (0.66-0.67) |
| Delirium-Coma | 0.71 (0.71-0.71) | 0.69 (0.69-0.69) | 0.70 (0.70-0.70) | 0.72 (0.72-0.73) |
| Any state-Death | 0.82 (0.82-0.82) | 0.83 (0.83-0.83) | 0.82 (0.82-0.82) | 0.77 (0.76-0.78) |